\documentclass[letterpaper]{article} % DO NOT CHANGE THIS
\usepackage{aaai25}  % DO NOT CHANGE THIS
\usepackage{times}  % DO NOT CHANGE THIS
\usepackage{helvet}  % DO NOT CHANGE THIS
\usepackage{courier}  % DO NOT CHANGE THIS
\usepackage[hyphens]{url}  % DO NOT CHANGE THIS
\usepackage{graphicx} % DO NOT CHANGE THIS
\usepackage{natbib}  % DO NOT CHANGE THIS AND DO NOT ADD ANY OPTIONS TO IT
\usepackage{caption} % DO NOT CHANGE THIS AND DO NOT ADD ANY OPTIONS TO IT
\usepackage{algorithm}
\usepackage{algorithmic}
\usepackage{amsmath} 
\usepackage{enumitem}
\usepackage{amsfonts}

\usepackage{newfloat}
\usepackage{listings}
\DeclareCaptionStyle{ruled}{labelfont=normalfont,labelsep=colon,strut=off} % DO NOT CHANGE THIS
\floatstyle{ruled}
\newfloat{listing}{tb}{lst}{}
\floatname{listing}{Listing}
\title{Fake It Till You Make It: Using Synthetic Data and  Domain Knowledge\\ for Improved Text-Based Learning for LGE Detection}

\author{
    Athira J Jacob \textsuperscript{\rm 1,2},
    Puneet Sharma \textsuperscript{\rm 1},
    Daniel Rueckert \textsuperscript{\rm 2,3}
}

\affiliations{
    \textsuperscript{\rm 1}Digital Technology and Innovation, Siemens Healthineers, Princeton, NJ, USA\\
    \textsuperscript{\rm 2} AI in Healthcare and Medicine, Klinikum rechts der Isar, Technical University of Munich, Germany\\
    \textsuperscript{\rm 3} Department of Computing, Imperial College London, UK
}

\begin{document}

\maketitle

\begin{abstract}
Detection of hyperenhancement from cardiac LGE MRI images is a complex task requiring significant clinical expertise. Although deep learning-based models have shown promising results for the task, they require large amounts of data with fine-grained annotations. Clinical reports generated for cardiac MR studies contain rich, clinically relevant information, including the location, extent and etiology of any scars present. Although recently developed CLIP-based training enables pretraining models with image-text pairs, it requires large amounts of data and further finetuning strategies on downstream tasks. In this study, we use various strategies rooted in domain knowledge to train a model for LGE detection solely using text from clinical reports, on a relatively small clinical cohort of 965 patients.  We improve performance through the use of synthetic data augmentation, by systematically creating scar images and associated text. In addition, we standardize the orientation of the images in an anatomy-informed way to enable better alignment of spatial and text features. We also use a captioning loss to enable fine-grained supervision and explore the effect of pretraining of the vision encoder on performance. Finally, ablation studies are carried out to elucidate the contributions of each design component to the overall performance of the model.     
\end{abstract}


\section{Introduction}

Late Gadolinium Enhancement (LGE) imaging—also referred to as Delayed Enhancement Imaging—plays a critical role in assessing myocardial viability. By highlighting affected areas with increased contrast uptake, it allows non-invasive detection and assessment of myocardial infarction, as well as ischemic and non-ischemic cardiomyopathies and other cardiac pathologies \cite{jenista2023revisiting}. However, detecting areas of hyperenhancement from LGE images (henceforth referred to as LGE or scar detection) is a challenging task. Enhancement can be subtle and is often influenced by spatial and temporal variations across different scanners, sequences, and study protocols. Moreover, image noise, artifacts, and varying LGE patterns add to the difficulty. These complexities make developing robust, generalizable automated solutions for LGE detection particularly challenging. Although deep learning (DL) approaches have achieved impressive performance in a range of medical imaging applications, they are heavily dependent on access to large, good quality, annotated datasets for increased accuracy and generalization. The annotation of LGE data, which requires precise delineation of both myocardial and enhancement regions, is especially complex due to its variability and requires substantial clinical expertise. These demands pose a limitation on large-scale DL training efforts in this area.

Meanwhile, clinical reports present a source of valuable information about LGE assessment. The reports are created from the cardiac magnetic resonance (CMR) study and contain clinician-written notes about the LGE, including location, extent, etiology, and other impressions. While these could be manually converted to binary labels to train a DL model, that is impractical for large amounts of data. Moreover, translating some of the information into discrete labels could potentially be a non-intuitive task. In such a situation, training directly with text offers an attractive alternative. 

Recently, Contrastive Language Image Pre-training (CLIP) \cite{radford2021learning} has achieved notable success in incorporating text supervision into vision models, for a wide range of downstream tasks in natural image processing, such as classification \cite{zhou2022learning},  object detection\cite{lin2023gridclip}, and segmentation \cite{luo2023segclip}. CLIP's approach aligns image and corresponding textual description embeddings within a shared latent space, facilitating a unified representation. This training paradigm has also been applied within medical imaging \cite{zhao2023clip}. However, CLIP models often require large datasets—often millions of image-text pairs—to learn robust associations.  Collecting such data, especially in specialized fields like medical imaging, is challenging due to limited availability and high annotation costs. These data demands can limit accessibility. In addition, despite the impressive zero-shot or few-shot performance, fine-tuning with task-specific data is often required to reach state-of-the-art (SoTA) performance. Many studies explore adapting CLIP models to downstream tasks through prompt tuning and linear probing. However, these require further training stages for the model, and is dependent on the size of the fine-tuning dataset. 

In this study, we train a model for LGE detection on a clinical cohort of 965 patients. We use text supervision from their corresponding clinical reports, without any further fine-tuning with discrete labels. We use domain knowledge to systematically augment the limited dataset with synthetic scar images and text. Moreover, we normalize the orientation of the images in an anatomy-informed way to facilitate correspondence between image and text features. Additionally, we add captioning loss to provide more granular supervision from the text.  Furthermore, the model’s encoder is pre-trained on a related task involving LGE images.  We obtain an average balanced accuracy of 0.83 on the test set. Though the model is supervised with patient-level descriptions, it can produce slice-level predictions due to the training strategy, which aids in interpretability. We also conduct ablation studies to assess the impact of each design choice and explore the effect of alternative encoder initialization strategies on model performance.

\section{Related Work}

\subsubsection{Vision-Language Training in the medical domain.}
Many studies have explored building specialized models in the medical domain \cite{convirt, wang2022medclip, zhang2024mediclip}. BiomedCLIP \cite{zhang2023biomedclip}, a vision-language model (VLM) was trained on 15M image-text pairs from the biomedical domain. Adapting these models to downstream tasks has been explored in various ways: a) End to end fine-tuning \cite{convirt, ikezogwo2024quilt},  b) Prompt tuning \cite{coop2022, cocoop}, c) Linear probing \cite{radford2021learning, coop2022,shakeri2024few}.  These methods still require additional training stages on a labeled "support" dataset for the downstream task, and can underperform in very low data cases, especially with complex tasks.  
Recently, multi-modal language models (MLLMs) have emerged as generalist models for various tasks \cite{medflamingo,medpalm,llavamed}. These are generative models capable of generating and processing long-form text. For instance, Med-Flamingo \cite{medflamingo} is an open-source, multimodal few-shot learner adapted to the medical domain. These models demonstrate impressive performance across a wide variety of tasks, such as visual question answering, classification, report generation, etc., on a wide range of modalities.

\subsubsection{Synthetic Data Generation.}

Augmentation with synthetic data has the potential to mitigate the issue of limited domain data in medical imaging. However, generating high-quality synthetic medical images remains challenging. Clip-Medfake \cite{medfake} leverages Stable Diffusion \cite{stablediffusion} to generate synthetic images, which are then used to pretrain a CLIP model before fine-tuning on actual medical data. Latte-CLIP \cite{cao2024latteclip} uses MLLMs to generate descriptive text for domain-specific images, which are then used for CLIP model fine-tuning. Data synthesis in these methods is done by pre-trained models or models trained from a small portion of the original training data.  They lack fine grained control over the synthesis process. CtrlSynth \cite{cao2024ctrlsynth} utilizes pre-trained models to identify concepts within an image, systematically alter these attributes, and generate synthetic text with LLMs. This synthetic text is then converted into images through Stable Diffusion, creating a comprehensive synthetic dataset that supports model training on controlled variations. 

\subsubsection{LGE Detection.}

There is no consensus regarding the optimal method for LGE analyses. Commonly used manual and semi-automatic methods clinically include manual planimetry, the Full Width Half Maximum (FWHM) approach \cite{amado2004accurate, hsu2006quantitative}, and n-std from remote myocardium. Comparative studies examining these methods \cite{flett2011evaluation, heiberg2022infarct} reveal significant variability in quantification results, highlighting issues with both reliability and reproducibility.

Recently, there have been many DL-based studies focused on automated scar detection from cardiac LGE images \cite{zhang2021cascaded, kim2024deep, girum2021automatic, yang2021hybrid}. For instance, Kim et al. \cite{kim2024deep} leveraged segmental information to train models that identify the presence or absence of LGE by transforming the images into polar coordinates based on the left ventricular (LV) center and the right ventricular (RV) insertion points. Additionally, several studies have explored infarct segmentation on the publicly available EMIDEC dataset \cite{lalande2020emidec} containing 150 patients (100 for training), achieving classification accuracies as high as 0.92 \cite{lalande2022deep} and a Dice coefficient of up to 0.71 for infarct segmentation \cite{zhang2021cascaded}. These models are trained on dense pixel-wise annotations of myocardial and scar tissue provided by clinical experts, thereby enhancing their ability to accurately segment scar regions.

\section{Methodology}

\subsubsection{Preliminaries: CLIP-based training}

CLIP training framework consists of a vision encoder and a text encoder to extract features from pairs of images and text, respectively. Each encoder is followed by a set of projection layers to project the features into a common embedding space. More specifically, the input image $x_{img}$ is encoded by the encoder $E_{img}$ into feature vector $f_{img} \in \mathbb{R}^n$. A projection module $P_{img}$   maps the features into the embedding $v \in \mathbb{R}^p$.

\[ v = P_{img}(E_{img}(x_{img}))\]
 
Similarly, the text encoder $E_{txt}$ encodes the input text into feature vector $f_{txt} \in \mathbb{R}^m$. A projection module $P_{txt}$   maps the features into the embedding $t \in \mathbb{R}^p$ .
\[ t = P_{txt}(E_{txt}(x_{txt}))\]

The similarity score is calculated using dot product as \[s = v_n \cdot t_n \] where $v_n, t_n$ represent L2 normalized vectors.
Then, cross-entropy loss (CE) is used to maximize the similarity score within the same pair, and minimize the same across pairs \cite{radford2021learning}.

\subsubsection{Preliminaries: Related tasks on LGE images.}

Prior to this study, we trained DL networks to segment the myocardium and detect anterior and posterior RV Insertion Points on LGE MRI Images. Ground truth (GT) annotations for these tasks are relatively easy to create. The myocardium segmentation network consisted of a UNet architecture, with DenseNet121 \cite{huang2017densely} encoder. It was trained on 7401 manually annotated images from 786 patients, from 3 centers. The model achieved an average Dice score of 0.88 on the test set of 131 patients (1197 images) from the 3 centers. The landmark detection model consisted of a UNet architecture, with ResNet18 \cite{resnet} encoder. It was trained on 3478 manually annotated images from 377 patients, from a single center to predict heatmaps centered around the landmark points. It achieved an average error 3.1 and 3.3 mm for the anterior and inferior RV Insertion Points, respectively. 

\subsection{Proposed Method}

\begin{figure}[t]
\centering
\includegraphics[width=0.95\columnwidth]{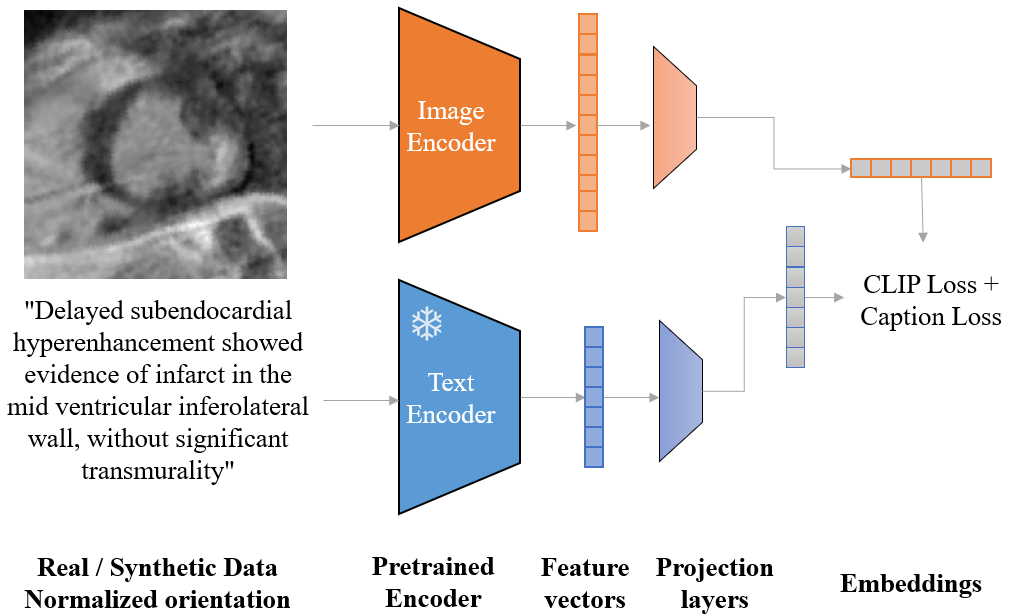} 
\caption{Overview of the proposed model}
\label{fig_model}
\end{figure}

We train a model to detect myocardial hyperenhancement from LGE images using relevant text from the clinical reports (Figure \ref{fig_model}). Due to the limited dataset size and the long-tailed distribution of LGE etiologies, we use domain knowledge to systematically augment the training data with synthetic image-text pairs. During training, the image-text pairs are aligned using global CLIP loss and a local caption loss. The image encoder is initialized with the weights from the myocardial segmentation network described in the previous section. Each of these parts is explained in detail in the following sections. During inference, we query the model using the following text: \texttt{there is hyperenhancement in the myocardium} and \texttt{there is no hyperenhancement in the myocardium}, denoting the positive and negative LGE classes respectively.

\subsection{Synthetic Data Generation}

\begin{figure}[t]
\centering
\includegraphics[width=0.99\columnwidth]{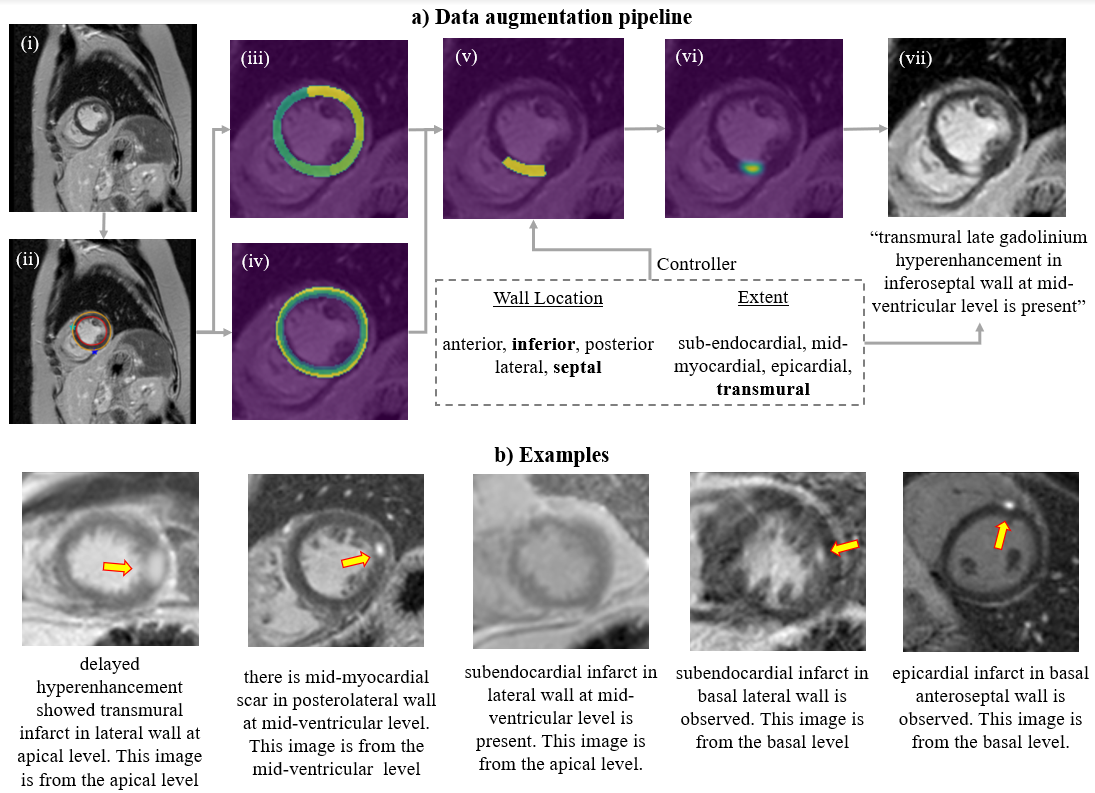} 
\caption{Synthetic Data Generation: a) Generation pipeline, b) Examples of synthetically generated images and corresponding captions. Text may refer to a region other than the one of the paired image (image 3). }
\label{fig_scar_pipeline}
\end{figure}

We add synthetic scars to real LGE images and create associated text descriptions (Figure \ref{fig_scar_pipeline}a). This allows us to augment the limited data, and systematically cover a wide variety of scar distributions. The scar is applied only to images with no prior LGE (as determined from the clinical report). A Controller module randomly chooses from a set of scar parameters. Then, a synthetic scar is added to the image, and text is created with these parameters. This is done at every training iteration, with a probability of $\lambda$. 

\subsubsection{Controller.} Wall location is randomly chosen as one of ["anterior", "inferior", "posterior"], or ["lateral", "septal"], or a combination of the words chosen from the two sets (Eg: "inferoseptal"). Scar extent is randomly chosen from ["sub-endocardial", "mid-myocardial", "epicardial", "transmural"]. "Transmural" signifies the scar spanning over more than 50\% of the myocardial thickness. 
\subsubsection{Image Generation.} For every negative LGE image (Figure 2a.i) at a given slice location, synthetic scars are added as follows:
\begin{enumerate}
\item Myocardial mask and anterior RVIP are determined using the previously trained DL networks (Figure 2a.ii).
\item Anterior RVIP is used to divide the myocardium into AHA segments: four if apical layer, or six segments if basal or mid (Figure 2a.iii).
\item The myocardial mask is divided equally into 3 concentric sections to represent endocardial, mid-myocardial and epicardial layers (Figure 2a.iv).
\item Wall location (chosen by the controller), along with slice location, is translated into AHA segments through hard-coded values. For eg, inferoseptal on basal level denotes segment 3.
\item A scar candidate region mask is created using an intersection between the identified AHA segments (Step 4) and chosen scar extent (by the controller). This is the "allowed" region for scar creation (Figure 2a.v).
\item Synthetic scar is created in the candidate region mask as a randomly placed, oriented and sized ellipse (Figure 2a.vi). A random pixel is chosen from the candidate region as the center of the synthetic scar. The radii of the ellipse are chosen randomly between set minimum and maximum values. The minimum and maximum values are determined as a fraction of the myocardial thickness at that point, depending on whether the scar is chosen to be transmural or within specific myocardial layers. The created ellipse is smoothed with a Gaussian filter for a more natural and continuous appearance.
\[
r_{min} = max(0.01, th * \rho_{min}) 
\]
\[
r_{max} = th * \rho_{max}  
\]
\[
\mathbf{r} = rand(r_{min}, r_{max}, 2)
\]
\[
\alpha = rand(0,\pi) 
\]
\[
\sigma = rand(0,1) * s_1 + s_2
\]

where, $th$ represents myocardial thickness at that point, $0 < \rho_{min}, \rho_{max} <= 1$ are hyperparameters representing ratios relative to myocardial thickness, $\mathbf{r} \in \mathbb{R}^2 $ are the radii of the major and minor axes of the ellipse, $\alpha$ represents the orientation of the major axis of the ellipse relative to the positive x-axis, and $\sigma$ represents the standard deviation of the Gaussian kernel used for smoothing. The created scar $M$ is min-max normalized to the range of $[0,1]$.
\item  The scar image $M$ is then blended with the image $I$ as, 
\[I_{synth} = I * (1-M) + \gamma * max(I) * M \]
\[ \gamma  = rand(b_1, b_2)\]
where $\gamma$ controls the brightness of the scar, and is randomly chosen between preset minimum and maximum values $b_1, b_2$ (Figure 2a.vii).
\end{enumerate}

\subsubsection{Text Generation.}
Given the chosen scar parameters (slice location, wall location, wall extent) from the controller module, the associated text description is synthesized using preset templates such as the following:\\
"\texttt{there is <extent>  delayed enhancement in <location> wall. This image is from <slice location> level.}"\\
or variations of this, where the words within \texttt{<>} are replaced with chosen scar parameters. The slice location of the input image is appended to stay consistent with real clinical text and contextualize the spatial location of the slice for the model (further explained in the Section  "Implementation Details: Text Encoder").  To add further variation to the text, the words "delayed enhancement", "delayed hyperenhancement", "late enhancement", "scar", "infarct" are used interchangeably. 
    
Examples of images with synthetically generated scar and corresponding text are shown in Figure \ref{fig_scar_pipeline}b.  

\subsection{Normalization of the LV orientation}

\begin{figure}[t]
\centering
\includegraphics[width=0.70\columnwidth]{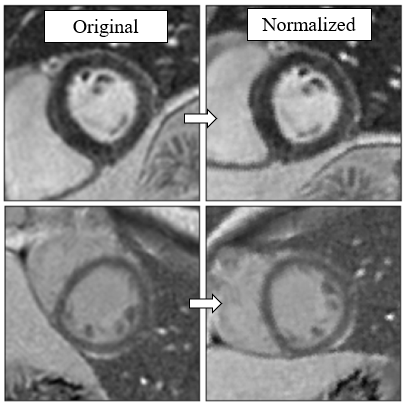} 
\caption{Anatomically informed normalization of LV orientation }
\label{fig_lv_norm}
\end{figure}

Clinical descriptions of LGE use words that describe location relative to the orientation of the LV, defined by the RV insertion point. The orientation of the LV can vary across patients, and even across images within the same patient. While this could potentially be learnt implicitly from a large cohort of image-caption pairs, this could prove to be a difficult challenge in a dataset of limited size. To help the model better associate position descriptors with image features, we standardize the orientation of the LV using the anterior and inferior RVIPs (Figure \ref{fig_lv_norm}). The RVIPs for each image are obtained from the landmark detection model described previously. Using the two insertion points, each image is rotated to position the line connecting them along the vertical axis of the image, with anterior RVIP being on the top.

\subsection{Caption loss}

CLIP loss aligns image and text embeddings on a global level. However, LGE descriptions from clinical reports are information-dense, with multiple words providing critical information about the location, extent and etiology of the scar. To encourage granular supervision on the level of the individual text tokens, we use a captioning loss similar to contrastive captioner models \cite{yu2022coca}. A multi-modal decoder is applied to the text tokens, consisting of layers of multi-headed, self-attention layers, followed by cross-attention layers attending to features from the vision encoder. The final layer is the classification layer that predicts the distribution of the next token over the supported vocabulary set.

\subsection{Task specific encoder}

We initialize the vision encoder with the weights of LGE myocardium segmentation model trained as described previously. Segmenting LGE myocardium is a closely related task to LGE detection, and is hypothesized to aid convergence. We later study the effect of this design choice by conducting ablation studies with various other image encoders. 

\section{Experiments}

\subsection{Data}

The data consists of 965 patients with cardiac MRI studies and clinical reports from a single center, of which 404 patients have reported LGE. The scans were performed on 1.5 T magnets (MAGNETOM Avanto, Siemens Healthcare, Erlangen, Germany) using a T1-weighted, phase sensitive inversion-recovery (PSIR), gradient-echo sequence, and were acquired 10 min after injection of a gadolinium-based contrast agent. The acquisition parameters are as follows, TR/TE: 2.4 /1.1 ms; flip angle: 50, slice thickness: 8 mm; in plane resolution between 1.5 × 1.5 6 $mm^2$ and 2.6 × 2.6 $mm^2$. The patients were divided in train, validation and testing splits of 772, 91 and 102 patients, respectively, while maintaining class distribution (LGE presence/absence).  

\subsection{Implementation Details}

\subsubsection{Vision Encoder.} The vision encoder consists of a Densenet121 \cite{huang2017densely} encoder with UNet decoder, with 5 downsampling layers. A MaxPool layer is added after the last layer of the DenseNet encoder to get a feature vector size of n = 1024.  The projection module consists of two Linear layers, separated by GelU non-linearity\cite{hendrycks2016gaussian} and followed by Dropout and LayerNorm \cite{ba2016layer}. 

For every patient, we select the segmented PSIR DICOM series  \cite{MUEHLBERG201813} using information in their DICOM tags, as this sequence theoretically has the higher spatial resolution required for LGE detection. In this dataset, these typically consist of 3 slices, covering apical, mid and basal regions of the heart. Each image is preprocessed according to the following steps: a) resizing to $1 mm \times 1 mm$ resolution, b) cropping to $112 \times 112$ dimension, centered around the LV. LV mask is obtained from the LGE segmentation network described previously, d) Upsampling $2\times$ to $224 \times 224$ e) capping intensities at the 98 percentile and f) normalizing to the range of [0,1].

\subsubsection{Text Encoder.} \label{section_text_encoding} We use the publicly available BiomedBERT \cite{zhang2023large}. Feature vector has size $m = 768$. The text encoder is held frozen for all experiments in this study. The projection module has the same architecture as described in the previous section and is optimized end-to-end during training. 

For each patient, text relevant to LGE imaging is extracted from the respective clinical report, from both the "Findings" and the "Impressions" sections, using a simple keyword search. The extracted text is split into individual sentences (henceforth also referred to as captions), from which one is sampled randomly for every iteration of training. 
However, this approach introduces an issue: the clinical text annotations are provided at the patient level, whereas the corresponding images represent specific heart sub-regions. To address this, we implement a straightforward solution by appending the phrase "\texttt{This image is from <slice location> level}" to each input text, where \texttt{<>} is replaced with basal, mid or apical, as per the image location. This is done consistently during both training and inference to help the model contextualize the image within the anatomical structure. While we limit our method here to three pre-selected slices for simplicity, this framework is adaptable to any number of images by adjusting the corresponding section tag.

\subsubsection{Scar augmentation parameters.}
In all experiments, we use $\lambda = 0.7$, $[\rho_{min}, \rho_{max}] = [0.1,0.4], [0.3,0.6], [0.7,0.1]$ for single-layer, two-layer, and transmural extents respectively, $s_1 = s_2 = 2 $, $b_1 = 0.8, b_2 = 1$. These were selected empirically to ensure image fidelity, anatomical relevance, and diversity across generated outputs.

\subsection{Baselines and Metrics}

We compare the proposed method against the following: 
\begin{enumerate} [label=\alph*)]
    \item BiomedCLIP \cite{zhang2023biomedclip}: We test the model on the same test set, using the same query text: \texttt{there is hyperenhancement in the myocardium} and \texttt{there is no hyperenhancement in the myocardium}. 
    \item MedFlamingo \cite{medflamingo}: The inference query is constructed as a prompt for few-shot, visual question-answering using examples: \texttt{<image 1> Does this cardiac LGE MR image show hyperenhancement in the LV myocardium? Answer: No. <image 2> Does this cardiac LGE MR image show hyperenhancement in the LV myocardium? Answer: Yes. <image 3>  Does this cardiac LGE MR image show hyperenhancement in the LV myocardium? Answer: }, where \texttt{image 1,image 2} are example images from the training set, without and with LGE respectively, and \texttt{image 3} is the query image. 
    \item Image-only classifier: A model is constructed to be identical to the CLIP based model, but without the text part, i.e the image encoder and a projection module followed by a classification head. The image encoder is initialized with the same pretrained weights. The classification head consists of two Linear layers separated by a GeLU non-linearity and a Dropout layer. The baseline model is trained with the GT binary labels extracted from the clinical reports, using binary cross entropy loss. 
\\\\
To reduce the stochastic variation in the results, multiple (n = 3) models were trained in every experiment and metrics were averaged. Balanced accuracy is adopted as the main metric to account for class imbalance in the dataset. 
Adam \cite{kingma2014adam}  optimizer is used with a learning rate of $1e^{-4} $. All models are trained across 4 x NVIDIA A100-SXM4-40GB GPUs using a batch size of 128, with fully distributed parallel processing.  Memory consumption (in GB) and throughput (in FPS) are measured on a single GPU of the same specifications. 
    
\end{enumerate}

\section{Results}

\begin{table}[]
\caption{Balanced accuracies, throughput and memory consumption of the compared methods (n = 102 patients)}
\label{main_results}
\begin{tabular}{l|c|c|c}
\hline
Method                     & FPS $\uparrow$  & Mem (GB) $\downarrow$ & Accuracy      \\ \hline \hline
\begin{tabular}[c]{@{}l@{}}BiomedCLIP\\ \cite{zhang2023biomedclip}\end{tabular}  & 74.4 & 0.76     & 0.58          \\
\begin{tabular}[c]{@{}l@{}}MedFlamingo\\ \cite{medflamingo}\end{tabular} & 1.4  & 31.09    & 0.50          \\
Image-only Classifier                                     & 84.5 & 0.04     & 0.77          \\ \hline
Proposed method                                           & 53.0 & 0.88     & \textbf{0.83} \\ \hline
\end{tabular}
\end{table}

Table 1 shows the quantitative results. The proposed method obtains a balanced accuracy of 0.83, outperforming the baselines. Both the publicly available medical VLMs (BiomedCLIP and MedFlamingo) encounter limitations on this task. These challenges likely stem from two key factors: (a) the VLMs were trained on a diverse array of medical imaging data, with cardiac MR constituting only a small subset, and LGE sequences representing an even smaller fraction of this subset; (b) LGE detection is an inherently challenging task that necessitates specialized clinical domain knowledge and the ability to analyze subtle, fine-grained features within highly localized regions of the images. The image-only classifier trained on this dataset demonstrates higher performance but lags behind the proposed method by 6 pp.

\begin{table}[]
\caption{Ablation studies: Model is retrained using leave-one-out strategy, keeping everything else the same. pp stands for percentage points.}
\label{tab:table2}
\begin{tabular}{p{5.3cm}|p{2.2cm}}
\hline
Method                           & Accuracy      \\ \hline \hline
Proposed                         & 0.83          \\ 
(-) Synthetic scar augmentation  & 0.73 (-10 pp) \\
(-) LV orientation normalization & 0.77 (-6 pp)  \\ 
(-) Caption loss                 & 0.80 (-3 pp)  \\ \hline
\end{tabular}
\end{table}

\subsubsection{Ablation Studies}

Table 2 illustrates the impact of omitting different components of the proposed method. Among the components, synthetic scar augmentation has the most significant impact on performance, followed by the normalization of LV orientation, and lastly, the caption loss.

Table 3 presents the impact of different initialization strategies for the vision encoder. As previously described, the proposed model uses a vision encoder pre-trained on the related task of myocardium segmentation in LGE images. Here, we explore two alternative initializations:
\begin{enumerate} [label=\alph*)]
\item Task agnostic encoder: This is a unimodal foundation model pre-trained on 36 million CMR images \cite{jacob2024towards}. It was trained in a self-supervised manner, without any labelled data, hence is agnostic to any specific task. The model consisted of a ViT-S architecture, and was pretrained across many diverse sequences of CMR, such as cine, LGE, and mapping.   
\item Imagenet pretrained encoder: This uses the same DenseNet-UNet architecture of the proposed model, but with the publicly available ImageNet trained weights.   
\end{enumerate}

Both initialization choices provide practical alternatives for when training data and/or labels for a directly related task is unavailable. Our results indicate that the task-agnostic CMR foundation model (FM), pre-trained in a self-supervised manner, outperforms the ImageNet pre-trained model; however, it still falls short compared to the model trained on a closely related task.

\begin{table}[]
\caption{Ablation study: Effect of different encoders on performance. }
\label{table3}
\begin{tabular}{p{5.5cm}|p{1.9cm}}
\hline
Method                           & Accuracy \\ \hline \hline
Proposed - Task specific encoder & \textbf{0.83}     \\
Task agnostic encoder - CMR FM \cite{jacob2024towards}   & 0.80     \\
Imagenet pretrained encoder      & 0.75     \\ \hline
\end{tabular}
\end{table}

\subsection{Qualitative Results}

\begin{figure}[t]
\centering
\includegraphics[width=0.95\columnwidth]{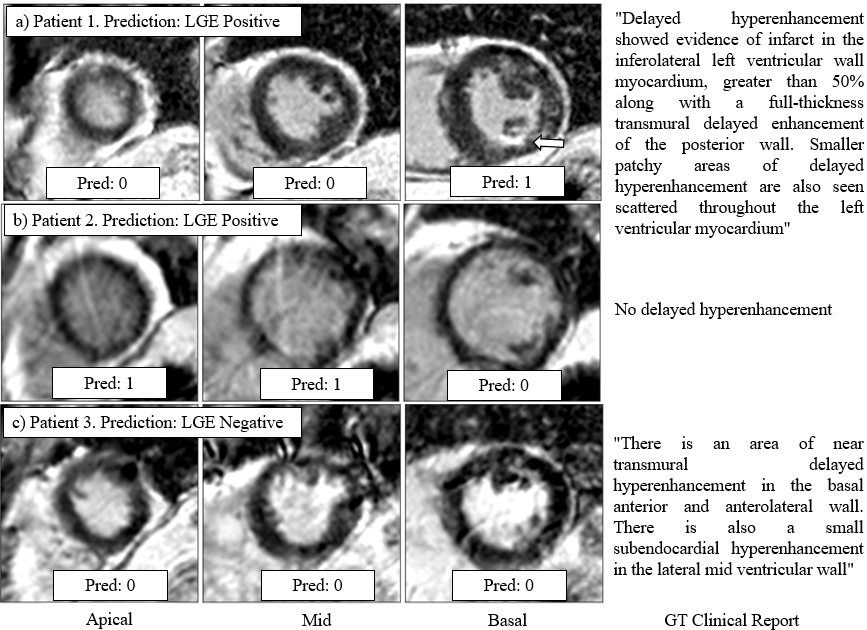} 
\caption{Qualitative results on real clinical data. Model predictions for each image, with the patient-level GT text}
\label{fig_examples}
\end{figure}

Fig \ref{fig_examples} visualizes the results for 3 patients. Patient 1 shows a true positive detection. Note that though the GT text describes LGE presence across the whole heart, the model is able to produce predictions for individual slices, which aids interpretability. Patient 2 presents a false positive, with the network predicting the LGE presence in apical and mid-ventricular slices. This could be because of the presence of streak artifacts in these two images, degrading image quality and possibly confounding the model. Patient 3 presents a false negative case, with no LGE detected by the model, despite the GT indicating LGE in basal slices. Visual inspection did not reveal LGE in these slices; however, further review of other LGE images in the study confirmed the presence of basal scarring at the level of the LV outflow tract. This highlights a method limitation, as selecting only three input images may omit critical details, reducing the model's efficacy. This is also observed in Patient 1, where the selected images do not reflect all of the described scar.

\section{Conclusions}

Text-based training using clinical reports offers an alternative to obtaining hard labels in the clinical domain, which might be expensive and challenging. However, typically used methods such as CLIP, require large amounts of pretraining data, and further finetuning stages for downstream tasks. We present a method that incorporates domain knowledge to enable CLIP based training for small datasets. We create synthetic image-text pairs to augment the training set, use anatomical information to normalize the orientation of the image, use additional caption loss to enable fine-grained supervision and use related-task pretraining to improve the accuracy for the task. We demonstrate the feasibility of text-based training for specific tasks on small datasets, followed by zero-shot inference without any further finetuning stages. 

\textbf{Disclaimer.} The concepts and information presented in this paper/presentation are based on research results that are not commercially available. Future commercial availability cannot be guaranteed.

% \subsection{Overlength Papers}
% If your paper is too long and you resort to formatting tricks to make it fit, it is quite likely that it will be returned to you. The best way to retain readability if the paper is overlength is to cut text, figures, or tables. There are a few acceptable ways to reduce paper size that don't affect readability. First, turn on \textbackslash frenchspacing, which will reduce the space after periods. Next, move all your figures and tables to the top of the page. Consider removing less important portions of a figure. If you use \textbackslash centering instead of \textbackslash begin\{center\} in your figure environment, you can also buy some space. For mathematical environments, you may reduce fontsize {\bf but not below 6.5 point}.

\bibliography{aaai25}

\begin{thebibliography}{40}
\providecommand{\natexlab}[1]{#1}

\bibitem[{Amado et~al.(2004)Amado, Gerber, Gupta, Rettmann, Szarf, Schock, Nasir, Kraitchman, and Lima}]{amado2004accurate}
Amado, L.~C.; Gerber, B.~L.; Gupta, S.~N.; Rettmann, D.~W.; Szarf, G.; Schock, R.; Nasir, K.; Kraitchman, D.~L.; and Lima, J.~A. 2004.
\newblock Accurate and objective infarct sizing by contrast-enhanced magnetic resonance imaging in a canine myocardial infarction model.
\newblock \emph{Journal of the American College of Cardiology}, 44(12): 2383--2389.

\bibitem[{Ba(2016)}]{ba2016layer}
Ba, J.~L. 2016.
\newblock Layer normalization.
\newblock \emph{arXiv preprint arXiv:1607.06450}.

\bibitem[{Cao et~al.(2024)Cao, Jaritz, Guillaumin, de~Charette, and Bazzani}]{cao2024latteclip}
Cao, A.-Q.; Jaritz, M.; Guillaumin, M.; de~Charette, R.; and Bazzani, L. 2024.
\newblock LatteCLIP: Unsupervised CLIP Fine-Tuning via LMM-Synthetic Texts.
\newblock \emph{arXiv preprint arXiv:2410.08211}.

\bibitem[{Cao, Najibi, and Mehta(2024)}]{cao2024ctrlsynth}
Cao, Q.; Najibi, M.; and Mehta, S. 2024.
\newblock CtrlSynth: Controllable Image Text Synthesis for Data-Efficient Multimodal Learning.
\newblock \emph{arXiv preprint arXiv:2410.11963}.

\bibitem[{Chen et~al.(2024)Chen, Zhao, Yue, Liu, Lv, Wang, and Zhou}]{medfake}
Chen, H.; Zhao, B.; Yue, G.; Liu, W.; Lv, C.; Wang, R.; and Zhou, F. 2024.
\newblock Clip-Medfake: Synthetic Data Augmentation With AI-Generated Content for Improved Medical Image Classification.
\newblock In \emph{2024 IEEE International Conference on Image Processing (ICIP)}, 3854--3860. IEEE.

\bibitem[{Flett et~al.(2011)Flett, Hasleton, Cook, Hausenloy, Quarta, Ariti, Muthurangu, and Moon}]{flett2011evaluation}
Flett, A.~S.; Hasleton, J.; Cook, C.; Hausenloy, D.; Quarta, G.; Ariti, C.; Muthurangu, V.; and Moon, J.~C. 2011.
\newblock Evaluation of techniques for the quantification of myocardial scar of differing etiology using cardiac magnetic resonance.
\newblock \emph{JACC: cardiovascular imaging}, 4(2): 150--156.

\bibitem[{Girum et~al.(2021)Girum, Skandarani, Hussain, Grayeli, Cr{\'e}hange, and Lalande}]{girum2021automatic}
Girum, K.~B.; Skandarani, Y.; Hussain, R.; Grayeli, A.~B.; Cr{\'e}hange, G.; and Lalande, A. 2021.
\newblock Automatic myocardial infarction evaluation from delayed-enhancement cardiac MRI using deep convolutional networks.
\newblock In \emph{Statistical Atlases and Computational Models of the Heart. M\&Ms and EMIDEC Challenges: 11th International Workshop, STACOM 2020, Held in Conjunction with MICCAI 2020, Lima, Peru, October 4, 2020, Revised Selected Papers 11}, 378--384. Springer.

\bibitem[{He et~al.(2016)He, Zhang, Ren, and Sun}]{resnet}
He, K.; Zhang, X.; Ren, S.; and Sun, J. 2016.
\newblock Deep residual learning for image recognition.
\newblock In \emph{Proceedings of the IEEE conference on computer vision and pattern recognition}, 770--778.

\bibitem[{Heiberg et~al.(2022)Heiberg, Engblom, Carlsson, Erlinge, Atar, Aletras, and Arheden}]{heiberg2022infarct}
Heiberg, E.; Engblom, H.; Carlsson, M.; Erlinge, D.; Atar, D.; Aletras, A.~H.; and Arheden, H. 2022.
\newblock Infarct quantification with cardiovascular magnetic resonance using" standard deviation from remote" is unreliable: validation in multi-centre multi-vendor data.
\newblock \emph{Journal of Cardiovascular Magnetic Resonance}, 24(1): 53.

\bibitem[{Hendrycks and Gimpel(2016)}]{hendrycks2016gaussian}
Hendrycks, D.; and Gimpel, K. 2016.
\newblock Gaussian error linear units (gelus).
\newblock \emph{arXiv preprint arXiv:1606.08415}.

\bibitem[{Hsu et~al.(2006)Hsu, Natanzon, Kellman, Hirsch, Aletras, and Arai}]{hsu2006quantitative}
Hsu, L.-Y.; Natanzon, A.; Kellman, P.; Hirsch, G.~A.; Aletras, A.~H.; and Arai, A.~E. 2006.
\newblock Quantitative myocardial infarction on delayed enhancement MRI. Part I: Animal validation of an automated feature analysis and combined thresholding infarct sizing algorithm.
\newblock \emph{Journal of Magnetic Resonance Imaging: An Official Journal of the International Society for Magnetic Resonance in Medicine}, 23(3): 298--308.

\bibitem[{Huang et~al.(2017)Huang, Liu, Van Der~Maaten, and Weinberger}]{huang2017densely}
Huang, G.; Liu, Z.; Van Der~Maaten, L.; and Weinberger, K.~Q. 2017.
\newblock Densely connected convolutional networks.
\newblock In \emph{Proceedings of the IEEE conference on computer vision and pattern recognition}, 4700--4708.

\bibitem[{Ikezogwo et~al.(2024)Ikezogwo, Seyfioglu, Ghezloo, Geva, Sheikh~Mohammed, Anand, Krishna, and Shapiro}]{ikezogwo2024quilt}
Ikezogwo, W.; Seyfioglu, S.; Ghezloo, F.; Geva, D.; Sheikh~Mohammed, F.; Anand, P.~K.; Krishna, R.; and Shapiro, L. 2024.
\newblock Quilt-1m: One million image-text pairs for histopathology.
\newblock \emph{Advances in neural information processing systems}, 36.

\bibitem[{Jacob et~al.(2024)Jacob, Borgohain, Chitiboi, Sharma, Comaniciu, and Rueckert}]{jacob2024towards}
Jacob, A.~J.; Borgohain, I.; Chitiboi, T.; Sharma, P.; Comaniciu, D.; and Rueckert, D. 2024.
\newblock Towards a vision foundation model for comprehensive assessment of Cardiac MRI.
\newblock \emph{arXiv preprint arXiv:2410.01665}.

\bibitem[{Jenista et~al.(2023)Jenista, Wendell, Azevedo, Klem, Judd, Kim, and Kim}]{jenista2023revisiting}
Jenista, E.~R.; Wendell, D.~C.; Azevedo, C.~F.; Klem, I.; Judd, R.~M.; Kim, R.~J.; and Kim, H.~W. 2023.
\newblock Revisiting how we perform late gadolinium enhancement CMR: insights gleaned over 25 years of clinical practice.
\newblock \emph{Journal of Cardiovascular Magnetic Resonance}, 25(1): 18.

\bibitem[{Kim, Chung, and Choe(2024)}]{kim2024deep}
Kim, Y.-C.; Chung, Y.; and Choe, Y.~H. 2024.
\newblock Deep learning for classification of late gadolinium enhancement lesions based on the 16-segment left ventricular model.
\newblock \emph{Physica Medica}, 117: 103193.

\bibitem[{Kingma(2014)}]{kingma2014adam}
Kingma, D.~P. 2014.
\newblock Adam: A method for stochastic optimization.
\newblock \emph{arXiv preprint arXiv:1412.6980}.

\bibitem[{Lalande et~al.(2020)Lalande, Chen, Decourselle, Qayyum, Pommier, Lorgis, de~la Rosa, Cochet, Cottin, Ginhac et~al.}]{lalande2020emidec}
Lalande, A.; Chen, Z.; Decourselle, T.; Qayyum, A.; Pommier, T.; Lorgis, L.; de~la Rosa, E.; Cochet, A.; Cottin, Y.; Ginhac, D.; et~al. 2020.
\newblock Emidec: a database usable for the automatic evaluation of myocardial infarction from delayed-enhancement cardiac MRI.
\newblock \emph{Data}, 5(4): 89.

\bibitem[{Lalande et~al.(2022)Lalande, Chen, Pommier, Decourselle, Qayyum, Salomon, Ginhac, Skandarani, Boucher, Brahim et~al.}]{lalande2022deep}
Lalande, A.; Chen, Z.; Pommier, T.; Decourselle, T.; Qayyum, A.; Salomon, M.; Ginhac, D.; Skandarani, Y.; Boucher, A.; Brahim, K.; et~al. 2022.
\newblock Deep learning methods for automatic evaluation of delayed enhancement-MRI. The results of the EMIDEC challenge.
\newblock \emph{Medical Image Analysis}, 79: 102428.

\bibitem[{Li et~al.(2024)Li, Wong, Zhang, Usuyama, Liu, Yang, Naumann, Poon, and Gao}]{llavamed}
Li, C.; Wong, C.; Zhang, S.; Usuyama, N.; Liu, H.; Yang, J.; Naumann, T.; Poon, H.; and Gao, J. 2024.
\newblock Llava-med: Training a large language-and-vision assistant for biomedicine in one day.
\newblock \emph{Advances in Neural Information Processing Systems}, 36.

\bibitem[{Lin and Gong(2023)}]{lin2023gridclip}
Lin, J.; and Gong, S. 2023.
\newblock Gridclip: One-stage object detection by grid-level clip representation learning.
\newblock \emph{arXiv preprint arXiv:2303.09252}.

\bibitem[{Luo et~al.(2023)Luo, Bao, Wu, He, and Li}]{luo2023segclip}
Luo, H.; Bao, J.; Wu, Y.; He, X.; and Li, T. 2023.
\newblock Segclip: Patch aggregation with learnable centers for open-vocabulary semantic segmentation.
\newblock In \emph{International Conference on Machine Learning}, 23033--23044. PMLR.

\bibitem[{Moor et~al.(2023)Moor, Huang, Wu, Yasunaga, Dalmia, Leskovec, Zakka, Reis, and Rajpurkar}]{medflamingo}
Moor, M.; Huang, Q.; Wu, S.; Yasunaga, M.; Dalmia, Y.; Leskovec, J.; Zakka, C.; Reis, E.~P.; and Rajpurkar, P. 2023.
\newblock Med-flamingo: a multimodal medical few-shot learner.
\newblock In \emph{Machine Learning for Health (ML4H)}, 353--367. PMLR.

\bibitem[{Muehlberg et~al.(2018)Muehlberg, Arnhold, Fritschi, Funk, Prothmann, Kermer, Zange, {von Knobelsdorff-Brenkenhoff}, and Schulz-Menger}]{MUEHLBERG201813}
Muehlberg, F.; Arnhold, K.; Fritschi, S.; Funk, S.; Prothmann, M.; Kermer, J.; Zange, L.; {von Knobelsdorff-Brenkenhoff}, F.; and Schulz-Menger, J. 2018.
\newblock Comparison of fast multi-slice and standard segmented techniques for detection of late gadolinium enhancement in ischemic and non-ischemic cardiomyopathy – a prospective clinical cardiovascular magnetic resonance trial.
\newblock \emph{Journal of Cardiovascular Magnetic Resonance}, 20(1): 13.

\bibitem[{Radford et~al.(2021)Radford, Kim, Hallacy, Ramesh, Goh, Agarwal, Sastry, Askell, Mishkin, Clark et~al.}]{radford2021learning}
Radford, A.; Kim, J.~W.; Hallacy, C.; Ramesh, A.; Goh, G.; Agarwal, S.; Sastry, G.; Askell, A.; Mishkin, P.; Clark, J.; et~al. 2021.
\newblock Learning transferable visual models from natural language supervision.
\newblock In \emph{International conference on machine learning}, 8748--8763. PMLR.

\bibitem[{Rombach et~al.(2022)Rombach, Blattmann, Lorenz, Esser, and Ommer}]{stablediffusion}
Rombach, R.; Blattmann, A.; Lorenz, D.; Esser, P.; and Ommer, B. 2022.
\newblock High-resolution image synthesis with latent diffusion models.
\newblock In \emph{Proceedings of the IEEE/CVF conference on computer vision and pattern recognition}, 10684--10695.

\bibitem[{Shakeri et~al.(2024)Shakeri, Huang, Silva-Rodr{\'\i}guez, Bahig, Tang, Dolz, and Ben~Ayed}]{shakeri2024few}
Shakeri, F.; Huang, Y.; Silva-Rodr{\'\i}guez, J.; Bahig, H.; Tang, A.; Dolz, J.; and Ben~Ayed, I. 2024.
\newblock Few-Shot Adaptation of Medical Vision-Language Models.
\newblock In \emph{International Conference on Medical Image Computing and Computer-Assisted Intervention}, 553--563. Springer.

\bibitem[{Singhal et~al.(2023)Singhal, Tu, Gottweis, Sayres, Wulczyn, Hou, Clark, Pfohl, Cole-Lewis, Neal et~al.}]{medpalm}
Singhal, K.; Tu, T.; Gottweis, J.; Sayres, R.; Wulczyn, E.; Hou, L.; Clark, K.; Pfohl, S.; Cole-Lewis, H.; Neal, D.; et~al. 2023.
\newblock Towards expert-level medical question answering with large language models.
\newblock \emph{arXiv preprint arXiv:2305.09617}.

\bibitem[{Wang et~al.(2022)Wang, Wu, Agarwal, and Sun}]{wang2022medclip}
Wang, Z.; Wu, Z.; Agarwal, D.; and Sun, J. 2022.
\newblock Medclip: Contrastive learning from unpaired medical images and text.
\newblock \emph{arXiv preprint arXiv:2210.10163}.

\bibitem[{Yang and Wang(2021)}]{yang2021hybrid}
Yang, S.; and Wang, X. 2021.
\newblock A hybrid network for automatic myocardial infarction segmentation in delayed enhancement-mri.
\newblock In \emph{Statistical Atlases and Computational Models of the Heart. M\&Ms and EMIDEC Challenges: 11th International Workshop, STACOM 2020, Held in Conjunction with MICCAI 2020, Lima, Peru, October 4, 2020, Revised Selected Papers 11}, 351--358. Springer.

\bibitem[{Yu et~al.(2022)Yu, Wang, Vasudevan, Yeung, Seyedhosseini, and Wu}]{yu2022coca}
Yu, J.; Wang, Z.; Vasudevan, V.; Yeung, L.; Seyedhosseini, M.; and Wu, Y. 2022.
\newblock Coca: Contrastive captioners are image-text foundation models.
\newblock \emph{arXiv preprint arXiv:2205.01917}.

\bibitem[{Zhang et~al.(2023{\natexlab{a}})Zhang, Xu, Usuyama, Bagga, Tinn, Preston, Rao, Wei, Valluri, Wong et~al.}]{zhang2023large}
Zhang, S.; Xu, Y.; Usuyama, N.; Bagga, J.; Tinn, R.; Preston, S.; Rao, R.; Wei, M.; Valluri, N.; Wong, C.; et~al. 2023{\natexlab{a}}.
\newblock Large-scale domain-specific pretraining for biomedical vision-language processing.
\newblock \emph{arXiv preprint arXiv:2303.00915}, 2(3): 6.

\bibitem[{Zhang et~al.(2023{\natexlab{b}})Zhang, Xu, Usuyama, Xu, Bagga, Tinn, Preston, Rao, Wei, Valluri et~al.}]{zhang2023biomedclip}
Zhang, S.; Xu, Y.; Usuyama, N.; Xu, H.; Bagga, J.; Tinn, R.; Preston, S.; Rao, R.; Wei, M.; Valluri, N.; et~al. 2023{\natexlab{b}}.
\newblock BiomedCLIP: a multimodal biomedical foundation model pretrained from fifteen million scientific image-text pairs.
\newblock \emph{arXiv preprint arXiv:2303.00915}.

\bibitem[{Zhang et~al.(2024)Zhang, Xu, Qiu, Yan, Lang, and Zhou}]{zhang2024mediclip}
Zhang, X.; Xu, M.; Qiu, D.; Yan, R.; Lang, N.; and Zhou, X. 2024.
\newblock Mediclip: Adapting clip for few-shot medical image anomaly detection.
\newblock In \emph{International Conference on Medical Image Computing and Computer-Assisted Intervention}, 458--468. Springer.

\bibitem[{Zhang(2021)}]{zhang2021cascaded}
Zhang, Y. 2021.
\newblock Cascaded convolutional neural network for automatic myocardial infarction segmentation from delayed-enhancement cardiac MRI.
\newblock In \emph{Statistical Atlases and Computational Models of the Heart. M\&Ms and EMIDEC Challenges: 11th International Workshop, STACOM 2020, Held in Conjunction with MICCAI 2020, Lima, Peru, October 4, 2020, Revised Selected Papers 11}, 328--333. Springer.

\bibitem[{Zhang et~al.(2022)Zhang, Jiang, Miura, Manning, and Langlotz}]{convirt}
Zhang, Y.; Jiang, H.; Miura, Y.; Manning, C.~D.; and Langlotz, C.~P. 2022.
\newblock Contrastive learning of medical visual representations from paired images and text.
\newblock In \emph{Machine Learning for Healthcare Conference}, 2--25. PMLR.

\bibitem[{Zhao et~al.(2023)Zhao, Liu, Wu, Wang, Li, Wang, Teng, Liu, Cui, Wang et~al.}]{zhao2023clip}
Zhao, Z.; Liu, Y.; Wu, H.; Wang, M.; Li, Y.; Wang, S.; Teng, L.; Liu, D.; Cui, Z.; Wang, Q.; et~al. 2023.
\newblock Clip in medical imaging: A comprehensive survey.
\newblock \emph{arXiv preprint arXiv:2312.07353}.

\bibitem[{Zhou et~al.(2022{\natexlab{a}})Zhou, Yang, Loy, and Liu}]{cocoop}
Zhou, K.; Yang, J.; Loy, C.~C.; and Liu, Z. 2022{\natexlab{a}}.
\newblock Conditional prompt learning for vision-language models.
\newblock In \emph{Proceedings of the IEEE/CVF conference on computer vision and pattern recognition}, 16816--16825.

\bibitem[{Zhou et~al.(2022{\natexlab{b}})Zhou, Yang, Loy, and Liu}]{zhou2022learning}
Zhou, K.; Yang, J.; Loy, C.~C.; and Liu, Z. 2022{\natexlab{b}}.
\newblock Learning to prompt for vision-language models.
\newblock \emph{International Journal of Computer Vision}, 130(9): 2337--2348.

\bibitem[{Zhou et~al.(2022{\natexlab{c}})Zhou, Yang, Loy, and Liu}]{coop2022}
Zhou, K.; Yang, J.; Loy, C.~C.; and Liu, Z. 2022{\natexlab{c}}.
\newblock Learning to prompt for vision-language models.
\newblock \emph{International Journal of Computer Vision}, 130(9): 2337--2348.

\end{thebibliography}

\end{document}